\documentclass[11pt]{article}

\usepackage[margin=1in]{geometry}  
\usepackage{setspace}  
\usepackage{caption}

\usepackage[fleqn]{amsmath}  
\usepackage{amssymb}
\usepackage{empheq} 
\usepackage{bm,upgreek}  

\usepackage{changepage}

\usepackage{url}

\usepackage{algorithmic,algorithm}  

\usepackage{graphicx}  
\usepackage{xspace}

\usepackage{color}  

\usepackage{tikz}
\usetikzlibrary{fit,positioning}
\usetikzlibrary{bayesnet}
\usepackage{wrapfig}

\usepackage{footmisc}
\newcounter{secnum}
\usepackage[all]{nowidow}

\usepackage{titlesec}
\titleformat*{\section}{\Large\bfseries}
\titleformat*{\subsection}{\onehalfspacing\Large}
\titleformat*{\subsubsection}{\large\bfseries}
\titleformat*{\paragraph}{\large\bfseries}
\titleformat*{\subparagraph}{\large\bfseries}

\usepackage{fancyhdr}
\usepackage{indentfirst}
\usepackage{lipsum}  

\usepackage{adjustbox}
\let\oldhash\#%
\DeclareRobustCommand{\#}{\adjustbox{valign=B,totalheight=.35\baselineskip}{\oldhash}}%

\newfont{\namefont}{cmr10 at 12.5pt}

\begin{document}
\vspace*{0cm}
\begin{center}
\noindent{\LARGE Foundations of Intelligence in Natural and Artificial Systems:\\[.25cm]  A Workshop Report}
\vspace{.5cm}

\rule{0.75\textwidth}{.4pt}
\vspace{.5cm}

\begin{minipage}{.3\textwidth}
\centering
\onehalfspacing
{\namefont Tyler Millhouse}\\ Santa Fe Institute\\ tyler.millhouse@santafe.edu\\
\end{minipage}
\begin{minipage}{.3\textwidth}
\centering
\onehalfspacing
{\namefont Melanie Moses}\\ University of New Mexico\\ melaniem@cs.unm.edu\\
\end{minipage}
\begin{minipage}{.3\textwidth}
\centering
\onehalfspacing
{\namefont Melanie Mitchell}\\ Santa Fe Institute\\ mm@santafe.edu\\
\end{minipage}\vspace{.5cm}

\rule{0.75\textwidth}{.4pt}
\vspace{.5cm}
\begin{quote}
\textbf{Abstract:} In March of 2021, the Santa Fe Institute hosted a workshop as part of its Foundations of Intelligence in Natural and Artificial Systems project. This project seeks to advance the field of artificial intelligence by promoting interdisciplinary research on the nature of intelligence. During the workshop, speakers from diverse disciplines gathered to develop a taxonomy of intelligence, articulating their own understanding of intelligence and how their research has furthered that understanding. In this report, we summarize the insights offered by each speaker and identify the themes that emerged during the talks and subsequent discussions. 
\end{quote}

\end{center}

\newpage
\tableofcontents
\newpage
\section{Overview}

\noindent The field of artificial intelligence is marked by a patchwork of astonishing successes and puzzling failures. AI programs have, one-by-one, dethroned human grandmasters in a number of deep and difficult games. At the same time, AI programs struggle with many tasks that are---to humans and other animals---little more than commonsense. This state of affairs is not an artifact of recent advances in isolated domains, but a longstanding problem in the field. In 1988, Hans Moravec famously noted that, ``It is comparatively easy to make computers exhibit adult level performance on intelligence tests or playing checkers, and difficult or impossible to give them the skills of a one-year-old when it comes to perception and mobility'' (p. 15). While considerable progress has since been made in these domains, the asymmetry described by Moravec has proven surprisingly resilient.

One way to explain this asymmetry is to note that our concept of intelligence is colored by our subjective experience of cognition. As Marvin Minsky observed, tasks involving effortful thought are too often believed to require more intelligence than tasks that we perform easily, automatically, or even unconsciously (1986). For this reason, humans tend to think that calculus is hard but riding a bike or appreciating a joke is easy. The unanticipated difficulty of these problems as well as our basic misconceptions about intelligence itself contribute to the persisting asymmetry of progress in AI. Without an appreciation of what problems require intelligence and what intelligence is, it is harder to develop appropriate metrics to evaluate intelligence and to guide intelligence research.

Of course, artificial intelligence is but one of many fields devoted to the study of intelligence. The other cognitive sciences (including psychology, neuroscience, and philosophy of mind) are deeply concerned with the foundations of intelligence---both natural and artificial. In addition, many fields outside the cognitive sciences have important and relevant insights on the nature of intelligence (including fields such as biology, ethology, and sociology). To refine our concept of intelligence, it is essential to examine intelligence in all its forms---whether human, animal, machine, or collective. This, in turn, requires experts in AI and cognitive science to critically engage with fields outside their traditional canon.
 
A central goal of this workshop was to gather researchers from many disciplines to consider foundational questions such as: How do you define ``intelligence'' or ``intelligences''?  What features of behavior do you think of as requiring an idea of intelligence? What are the most important questions about intelligence or cognition more broadly that you think about in your domain of study? What approaches could best provide an answer to these questions? As these questions illustrate, the workshop aimed to further \textit{a taxonomy of intelligence across disciplines}, a taxonomy that helps us to appreciate how different fields understand intelligence, what those fields have learned, and what questions remain open. In what follows, we summarize the key ideas presented and reflect upon their broader implications for our understanding of intelligence. 

The workshop was funded by a grant from the National Science Foundation (\#2020103) as part of the Foundations of Intelligence in Natural and Artificial Systems project at the Santa Fe Institute. Leading this project are Melanie Mitchell (PI) and Melanie Moses (Co-PI). The central aim of the grant project is to plan for the creation of an AI Institute at SFI. A key part of the planning process involves building collaborative relationships between SFI faculty and other researchers working in five key areas. These areas represent five broad themes of research that naturally incorporate scholars from multiple fields. This workshop aimed to bring together experts whose research can inform a taxonomy of intelligence across disciplines. Future workshops will focus on the four remaining themes, including the development and life history of intelligence, collective intelligence, the role of abstraction and analogy in intelligence, and evolutionary intelligence.  

\section{Summaries of Talks and Discussions}

\subsection[``Why AI is Harder Than We Think'' (Melanie Mitchell)]{``Why AI is Harder Than We Think''}

\begin{adjustwidth}{1cm}{1cm}
\onehalfspacing
\textit{Melanie Mitchell is the principal investigator for the Foundations of Intelligence in Natural and Artificial Systems project, the Davis Professor of Complexity at the Santa Fe Institute, and Professor of Computer Science at Portland State University.}
\end{adjustwidth}

\noindent In the opening talk of the workshop, Melanie Mitchell outlines five fallacies that frequently arise when thinking about progress in artificial intelligence. The first fallacy is that \textit{narrow AI is on a continuum with general AI}. The idea here is that progress in a particular domain of AI is ultimately progress toward domain general AI. The assumption, then, is a continuity between the kind of intelligence exhibited by, say, a chess-playing computer and the kind of intelligence a human being exhibits in everyday life. On the contrary, Mitchell argues that there may be significant differences between these intelligences, and that advances in the former may not apply to the latter.

The second fallacy is the tendency to think that things that are subjectively difficult \textit{for us} are difficult in some objective sense and things that are subjectively easy \textit{for us} are easy in some objective sense. This thinking leads us to believe that playing chess or Go should be difficult for AI systems and that perception and common sense should be comparatively easy. If this were true, Mitchell argues, then it is difficult to explain why computers can beat grandmasters at chess, while young children can beat computers at common sense reasoning tasks. 

The third fallacy is the tendency of researchers to uncritically accept the ``wishful mnemonics'' used to describe AI performance (McDermott, 1976, p. 4). For example, when we speak of an AI program that extracts information from texts, we might find it natural to say that the AI program is ``reading'' the texts. Worse, we may find it natural to believe--without closer examination--that the AI's activity is closely analogous to human reading. While thinking in terms of analogies is entirely natural and, on Mitchell's view, central to intelligence, she here identifies a failure mode for this type of reasoning that often results in our ``over attributing'' abilities to AI. 
 
The fourth fallacy is that intelligence and learning are best understood as phenomena going on exclusively inside the brain. The problem here is not that there is something over and above the brain responsible for intelligence, but rather that it is difficult to understand these phenomena unless we view the brain both as embodied and as situated in a particular environmental context. This fallacy leads us to believe that machine learning can extract commonsense understanding from data without substantial interaction with the world. On the contrary, work in developmental psychology suggests that human learning (which outperforms machine learning in many respects) is deeply dependent on this kind of interaction. 

The fifth fallacy is that intelligence can be characterized ``purely''---without considering the limitations, biases, strategies, or goals of any particular kind of intelligent agent. On the contrary, it is clear that (at least) our form of intelligence is deeply connected to these factors. For example, human learning and intelligence depends on broad motivations (like curiosity), specific learning strategies (like gaze-monitoring), and particular objectives (like cooperation and imitation). It would be too hasty to assume that these are mere design quirks of human beings rather than convergent design features required (to some degree) in any generally intelligent system. 

In framing these points, Mitchell outlined her own tentative view of intelligence. Intelligence involves a simulatable model of oneself and one's environment that generates predictions and expectations. Further, it is closely coupled to the ability to affect the world---whether by manipulating physical objects or by altering the mental states of others. Finally, it centrally involves the ability to abstract from particular cases and apply lessons learned in one context to relevantly analogous contexts. 

\subsection[``Knowing What To Do When You Don't Know What To Do'' (Daniel Dennett)]{``Knowing What To Do When You Don't Know What To Do''}

\begin{adjustwidth}{1cm}{1cm}
\onehalfspacing
\textit{Daniel Dennett is the Director of the Center for Cognitive Studies, University Professor, and the Austin B. Fletcher Professor of Philosophy at Tufts University and an External Professor at the Santa Fe Institute.}
\end{adjustwidth}

Daniel Dennett's talk focused on two key themes. First, he discussed the nature of intelligence in agents. Second, he discussed the risks of creating artificially intelligent agents and how those risks might be mitigated. Dennett begins by distinguishing intelligence from competence. He notes that many animals (even insect larvae) display competence with tasks whose purpose and rationale they do not comprehend. In contrast, a human performing a similar task might or might not exhibit greater competence, but they \textit{would} have the capacity to understand what they are doing and why. For Dennett, the latter is better understood as an example of intelligence. 

To capture the difference between these cases, Dennett proposes that intelligence is the capacity to ``know what to do when you don't know what to do.'' This distinguishes the ability to call forth a particular competence when appropriate, from the ability to reason about how best to solve a particular problem. Note that, on Dennett's definition, merely deploying a problem-solving heuristic is not enough. Rather, real agents face problems that cannot be solved by any single heuristic, and intelligence is the capacity to choose wisely among a large array of available heuristics.

Of course, this higher-order capacity is not as rapid or as automatic as reflexively deploying a single heuristic, but it is more robust to the failure modes associated with such strategies. For this reason, it will often be reasonable to ``oversimplify and self-monitor'' (Dennett, 2008, p. 1). In other words, when faced with a particular type of problem, it often makes sense for agents to automatically deploy a single heuristic---\textit{provided} the agent monitors for relevant failure modes and (when necessary) selects a different strategy. 

This seems to suggest that we should create artificially intelligent agents that possess a wide range of heuristics as well as the ability to self-monitor and re-direct their efforts. Dennett urges us to resist this conclusion. Agents, he argues, possess convergent instrumental reasons to deceive/withhold information and to secure their own autonomy, and these motivations present serious barriers to the effective control of AI. Instead, he suggests that humans adopt a particular division of labor with AI systems. AI systems will generate candidate solutions/strategies and humans will evaluate those candidates and select among them. As Dennett argues, this not only allows us to use AI systems to substantially extend our human capacities, but also to mitigate the risks associated with AI.  

One challenge to this strategy is that we typically think of ourselves and our fellow humans as agents who consciously reason about how to accomplish goals given what we know about the world. Taking this \textit{intentional stance} towards ourselves and others is something that we find extremely natural, and it is reinforced by our social interactions. For example, parents and children frequently ask (and expect answers) to questions about why they are doing what they are doing (McGeer, 2007). As a result, we find it almost irresistible to attribute intentionality to any system that \textit{appears} to exhibit agency. This means that in order for humans to assume the proper role in Dennett's generate and test approach, AI researchers should be careful not to see anthropomorphism as a goal of AI design. The limitations of AI systems must be clear and well-understood in order to prevent human assessors from giving undue deference to AI systems.

\subsection[``Three Ages, Three Intelligences'' (Alison Gopnik)]{``Three Ages, Three Intelligences: Childhood, Adulthood, and Elderhood, and Exploration, Exploitation and Care''}

\begin{adjustwidth}{1cm}{1cm}
\onehalfspacing
\textit{Alison Gopnik is a Professor of Psychology and Affiliate Professor of Philosophy at the University of California at Berkeley.}
\end{adjustwidth}

\noindent Alison Gopnik discussed the connection between the distinctive life stages of humans and the types of intelligence that dominate these stages. A classic problem in machine learning is finding an appropriate trade-off between exploration and exploitation. This problem arises because once an agent discovers a beneficial behavioral strategy, deviating from that strategy in hopes of finding a better one means risking the benefits that could have been acquired by exploiting the original approach. Of course, there is also the possibility that our current best strategy is sub-optimal (perhaps significantly so) relative to untested alternatives. A common solution is to begin learning with a bias toward exploration that gradually shifts into a bias toward exploitation. 

Gopnik argues that humans have taken a similar approach via adaptive changes to their life history (Gopnik, et al., 2020). Long childhoods present an extended opportunity for exploratory learning before a period of adulthood where we put to use what we have learned. This approach is not exclusive to humans, she argues, but is reflected in a number of species where longer periods of immaturity are associated with higher adult intelligence. Further, studies have confirmed that children (and other juvenile animals) tend to engage in more exploratory learning than their adult counterparts. 

In addition to an extended period of childhood, human life history is also marked by menopause, which seems hard to rationalize in terms of personal fitness. Also, younger adults often devote resources to rearing children that are not their own (i.e., allo-parenting). Finally, pair-bonded humans care for their young with relatively high levels of paternal investment.  As Gopnik argues, grandparents, alloparents, and pair-bonded parents all enhance the value of our extended childhood by contributing to learning, with grandparents playing a special role. 

Younger adults are often burdened with securing vital resources, and they are often the most capable of doing so when compared to children and older adults. Older adults, of course, still possess vital knowledge that can be transmitted to children even if they are no longer as efficient at using this knowledge. For this reason, Gopnik argues that menopause and the tendency of grandparents to exhibit special care for their grandchildren are both part of an evolutionary strategy for maximizing the fitness of older adults whose physical abilities have diminished and the fitness of children whose abilities/opportunities for learning are at their greatest. The special care exhibited by grandparents is thus an important and distinctive element of the human learning process, with children specializing in exploration, younger adults specializing in exploitation, and older adults specializing in transmitting knowledge to developing children.

\subsection[``Designing Alien Intelligences'' (James Evans)]{``Designing Alien Intelligences: Programming Beyond the Limits of Collective Knowledge and Reason''}

\begin{adjustwidth}{1cm}{1cm}
\onehalfspacing
\textit{James Evans is Professor of Sociology and Director of the Knowledge Lab and Computational Social Science program at the University of Chicago and an External Professor at the Santa Fe Institute.}
\end{adjustwidth}

\noindent Building on themes raised by Daniel Dennett, James Evans considered how we might design AI systems that are specially tailored to augment the individual and collective  intelligence of human beings. Evans begins by noting the power of ensembles of machine learning models and of diverse groups of human researchers. He is particularly interested in the kind of information latent in the research output and social organization of human researchers. For example, recent work has shown that word embeddings learned from scientific papers support surprisingly accurate predictions about (as yet) unknown material properties (Tshitoyan, et al., 2019). Evans augments this kind of approach by considering not only the content of these papers but the connections (e.g., co-authorship) between their authors (Sourati \& Evans, 2021). This dramatically enhances (100\%-300\%) the accuracy of predictions about what material properties later research will identify. 

One striking observation about this model is that it fares better in predicting the inferences researchers will draw in the near term and fares worse at predicting which inferences researchers will draw in the more distant future. This suggests that the value of network information is reduced as research programs and social connections shift over time. Unfortunately, the value of predicting what human researchers will infer in the near term is somewhat limited. As Evans argues, it would be more valuable to augment the collective intelligence of human researchers by identifying inferences that are plausible, yet unlikely to be drawn given how researchers are connected and what they are working on. Not only can this help us to see blind spots in human research, it can give rise to an anti-crowd sourcing effect. In some cases, aggregating judgments of group members can yield better predictions than relying on any particular individual. However, it seems probable that many of the inferences that are ``close to'' inferences humans have already drawn are inferences humans have investigated but failed to confirm. Hence, it may be more profitable to explore inferences further afield. 

One worry for this kind of approach is that tuning AI systems to suggest ``alien'' inferences will undermine the scientific value of these inferences. Perhaps human researchers have few blind spots and generally tend to focus their efforts on nearly all of the most promising areas of research. On the contrary, Evans's research has shown that tuning AI systems to ``avoid the crowd'' still results in those systems generating promising hypotheses that human researchers are unlikely to generate themselves (Sourati \& Evans, 2021, p. 1).

In summary, Evans envisions AI as a kind of alien intelligence that (while not as adept at inference as human beings) can be tuned to explore the space of possible inferences in ways that human investigators are unlikely to pursue (Evans, 2020). Much like Daniel Dennett envisioned AI as a system for generating possible solutions which humans evaluate, Evans argues that AI can help to extend our human imagination by proposing (but not confirming) inferences that we are prone to miss. 

\subsection[``Tradition and the Individual Talent'' (David Krakauer)]{``Tradition and the Individual Talent: Collective Puzzling at the Limits of Individual Cognition''}

\begin{adjustwidth}{1cm}{1cm}
\onehalfspacing
\textit{David Krakauer is the President and William H. Miller Professor of Complex Systems at the Santa Fe Institute. }
\end{adjustwidth}

\noindent David Krakauer's talk explored a wide range of themes relating to the nature and study of intelligence. Beginning with a discussion of the cultural evolution of representation using the competitive Rubik's cube community as a model system. Krakauer argues that a number of phenomena central to individual and collective learning can be seen in the development and propagation of cube-solving techniques. For example, he suggests that we can think of the time scale of social learning as arising from the time scale of individual learning. He notes that as more advanced techniques are introduced, there is a learning curve for individuals to adopt and perfect the application of these techniques. This learning curve dictates the pace of collective advancement, resulting in a stair-step progression as new techniques are introduced and perfected.

Next, Krakauer turns to a more explicit analysis of how experts actually solve Rubik's cubes. The difficulty of this task and the appropriate method depends in large part on whether one is solving the cube in the sighted or blindfolded condition. The sighted condition not only reduces the memory demands of memorizing the cube, it allows decisions about the right moves to be made on the basis of a sparse (i.e., partial) representation of the cube's state. That said, individuals in the sighted condition have to memorize a more complicated set of rules for cube manipulation. Unlike memorizing the cube's state, however, individuals can commit these rules to long-term memory and practice in order to automate them. Perhaps unsurprisingly, error rates are lower in the sighted condition.

Krakauer argues that the ability to rely on a sparse cube representation allows for improved performance and, more generally, that how we represent a problem domain can either help or hurt us in finding useful rules and achieving good results. This suggests a more general account of intelligence. Krakauer holds that intelligence is the acquisition of representations/rules (e.g., from the community) that facilitate successful problem solving. Unlike other speakers, he is more prepared to call a system ``intelligent'' when it exhibits competence in a narrow domain and more willing to see the acquisition and implementation of specific rules as central to intelligence. The value of this approach is that is renders intelligence a Darwinian system allowing species-specific problem solving to be understood in their own terms. 

While this elicited some disagreement, Krakauer's account highlights a dimension of intelligence not emphasized by other speakers. Daniel Dennett, for example, spoke of less intelligent systems as exhibiting ``competence without comprehension'' and contrasted this with human intelligence. Krakauer, on the other hand, suggest that genuine intelligence is competence with a kind of ``learned incomprehension'' that comes from the acquisition of expertise. Champion tennis players, for example, are free to ponder higher-level game strategy precisely because they have automated the more mechanical aspects of play. This suggests a more complicated interplay between the kinds of competence at work in automatic/effortless cognition and the effortful cognition typically associated with intelligence. Krakauer calls this feature the ``Searle Inversion." 

\subsection[``Intelligence and the Nature of Embodiment'' (Ricard Sol\'{e})]{``Intelligence and the Nature of Embodiment: From Liquid to Solid''}

\begin{adjustwidth}{1cm}{1cm}
\onehalfspacing
\textit{Ricard Sol\'{e} is a Catalan Institute for Research and Advanced Studies Research Professor at Universitat Pompeu Fabra in Barcelona and an External Professor at the Santa Fe Institute.}
\end{adjustwidth}

\noindent Ricard Sol\'{e} considered the constraints that influence the evolutionary development of different kinds of intelligences. Sol\'{e} argues that we might gain traction on the nature of intelligence by identifying a space of possible intelligences and then exploring which parts of that space are occupied by \textit{actual} intelligences. More importantly, we should consider what kinds of constraints might give rise to patterns in the forms of intelligence represented in the world. For example, Sol\'{e} argues that mobile organisms (like ants) seem to be pushed towards relatively greater general intelligence by the vagaries inherent in finding one's way around a changing world. On the other hand, sessile organisms (like plants) seemed to be pushed toward a relatively low level of intelligence given the reduced value of intelligence in their niche. 

One area Sol\'{e} explores in detail is that of ``liquid brains.'' A common image of a brain is as a collection of fixed cells defined by their connections to other cells and how those connections change over time. Liquid brains are those which not only change their connections but also their structure. For example, groups of ants can (collectively) engage in surprisingly intelligent behavior, but the locations and connections between ants are not fixed. As the ants move about their environment, they form and break connections dynamically with their neighbors. However, this kind of flexibility, Sol\'{e} suggests, may impose a trade-off on the complexity of individual members of such collectives. He notes that species with larger colony sizes tend to have members that are (as individuals) more specialized and less intelligent. It is unclear to what extent this generalizes to other collectives, but it could tell us something quite important about the trade-offs inherent to developing a collective intelligence.  

\subsection[``How Other Minds Shape Our Own'' (Thalia Wheatley)]{``How Other Minds Shape Our Own: Intelligence as a Networked Phenomenon''}

\begin{adjustwidth}{1cm}{1cm}
\onehalfspacing
\textit{Thalia Wheatley is the Lincoln Filene Professor of Human Relations at Dartmouth College and an External Professor at the Santa Fe Institute.}
\end{adjustwidth}

\noindent Thalia Wheatley explored the effects of social relationships on cognition and how those effects manifest in the brain. Wheatley proposes that human social relationships can make us either more or less intelligent depending on the number and, more importantly, the type of connections we have with others. She outlines four key points supporting this hypothesis. First, we know that humans offload cognition to members of their social groups---e.g., by distributing skills and memories among group members. Second, we know that people differ in how they are connected to other members of their social network. For example, some individuals exhibit high connectivity in a general sense, being well-connected to people who are themselves well-connected. Other individuals, known as ``brokers,'' bridge the gaps between groups that are otherwise poorly connected. These connections are also highly salient to individual humans, and the patterns of brain activity produced when viewing another individual predict that individual's connectedness (Parkinson, Kleinbaum, \& Wheatley, 2017). 

Wheatley's third key point is that these connections shape how we think in profound and surprising ways. For example, across a diverse range of video clips, patterns of brain activity are more similar for more closely connected individuals, with friends exhibiting especially high levels of similarity (Parkinson, Kleinbaum, \& Wheatley, 2018). Further, when individuals converse about how to interpret a video clip, this enhances the synchronization of brain activity---both for the original clip and for novel clips (Sievers, et al., Preprint).  

Wheatley's fourth key point is that some people in our network are better at enhancing our cognitive abilities than others. One concern about the results just described is that synchronization can lead to a narrowing of thought about the issue at hand, and this might prevent individuals from discovering better solutions to the problems they face.  Fortunately, brokers may represent a potential fix for this problem by connecting social groups with diverse viewpoints. Brokers seem to possess special skills (learned by past experiences in diverse communities) that make them well-suited to bridging gaps between poorly connected social groups (Wood, Kleinbaum, \& Wheatley, 2019). This suggests an important role for both cognitive synchronization (e.g., in cognitive offloading) and for brokers who mitigate the insulating effects of this phenomenon and help to enhance our intelligence.  

\subsection[``Perception as Model Building'' (Bruno Olshausen)]{``Perception as Model Building''}

\begin{adjustwidth}{1cm}{1cm}
\onehalfspacing
\textit{Professor Bruno Olshausen is a Professor in the Helen Wills Neuroscience Institute, the School of Optometry at the University of California at Berkeley.  He also serves as Director of the Redwood Center for Theoretical Neuroscience}
\end{adjustwidth}

\noindent Bruno Olshausen examined the role of model building in perception and how these models are realized in the brain. Olshausen begins by outlining an intriguing method for probing perceptual models. Participants are shown an image whose contents are not easily recognizable, and they are asked to draw what they see. Once individuals have recognized the figure in the image, what they draw when prompted is radically different (and quite recognizable) compared to the abstract figures they previously drew (Dallenbach, 1951). This suggests that visual object recognition relies on building a model of what is depicted in a scene. In this respect, Olshausen echoes Melanie Mitchell's emphasis on modeling as a key component of intelligence.

This kind of model building is not restricted to human beings. Jumping spiders visually inspect their environment and plan their route toward a reward. They are not simply following a set of behavioral rules based on their immediate surroundings (Hill, 1979; Tarsitano \& Jackson, 1997). A key part of this kind of model building, Olshausen argues, is the ability to disentangle or \textit{factorize} different aspects of a scene. For example, humans struggle to perceive the actual color of light that falls on the retina. The reason for this is that the brain builds a model that disentangles the surface properties of an object from incident lighting. The result is that even when two parts of an image reflect the same color of light into the eye, they can be perceived as different colors based on the brain's model of how lighting is influencing the appearance of the surface. While this kind of factorization requires fairly strong constraints (e.g., priors) to operate effectively, Olshausen notes that even when we have isolated the relevant factors, the space of scenes which could be inferred is still staggering due to the combinatorial nature of the factors involved.

While currently we have very little direct evidence for how these models are realized in the brain, recent work revealing the neural circuit responsible for computing and representing head direction in the fly brain provides an important start.  Theoretical work had speculated that neurons recurrently connected in a ring topology could be used to produce an internal representation of an animal's heading in two-dimensional space (Zhang, 1996). Of course, having a ring of such connections does not require the cells to be physically arranged into a ring. Nevertheless, the brains of fruit flies contain a prominent doughnut-shaped structure (i.e., the ellipsoid body) which computes and represents heading (Seelig \& Jayaramen, 2015) presumably to economize on wiring lengths.  Importantly, this representation does not simply reflect the direct sensory input from any one modality because there are no sensors for heading.  Rather it must be built up by disentangling, or factorizing, the components within multiple sensory modalities (vision, vestibular, proprioceptive) containing information about heading apart from other factors.  In this way, it forms and explicitly represents an hypothesis regarding its place in the world, distinct from the particular format of sensory input, which can then be by used by other systems within the fly brain for navigation or other tasks.

In summary, this suggests that intelligence researchers should pay closer attention to both the role of disentanglement in modeling and the role of neural realization in understanding these models. Olshausen argues that existing approaches to AI (such as deep neural networks) may struggle precisely because they neglect these considerations. For example, neural networks do not engage in explicit processes of factorization nor do they involve a realistic representation of the organization or computation abilities of actual neurons. For these reasons, Olshausen is optimistic that approaches emphasizing disentanglement and neural realism will fare better in creating artificially intelligent systems. 

\subsection[``Understanding the Cognitive Brain'' (John Krakauer)]{``Understanding the Cognitive Brain''}

\begin{adjustwidth}{1cm}{1cm}
\onehalfspacing
\textit{John Krakauer is the John C. Malone Professor of Neurology, Neuroscience, and Physical Medicine \& Rehabilitation, the Director of the Brain, Learning, Animation, and Movement (BLAM) laboratory, and co-founder of the KATA project at Johns Hopkins University School of Medicine.}
\end{adjustwidth}

\noindent John Krakauer discussed the interplay between cognition and reflex, emphasizing the importance of what he calls ``intelligent reflexes'' (Krakauer, 2019, p. 825)  Krakauer begins by discussing the surprising degree of intelligence that can be involved in behaviors that are best thought of as reflexes. For example, he discusses early neuroscientific experiments that (among other things) involved decapitating frogs and determining what sorts of reflexes remain intact (Klein, 2017). One such reflex was using the leg to wipe irritants off the skin. Not only did this show a remarkable degree of sophistication for a reflex, the frog dutifully switched to the other leg when the first leg was amputated. Krakauer argues that this shows the degree to which reflexive control policies can underlie behavior we might initially believe require higher-level reasoning (e.g., modeling, simulation, and prediction). 

Much like David Krakauer, John Krakauer sees a central role for the automation of learned skills in intelligent behavior. Rather than undermining the importance of deliberate or effortful cognition, he sees this kind of cognition as essential to our ability to learn these automated behaviors. The key idea is that we have the ability to transform propositional knowledge into automatic, goal-directed responses (i.e., ``intelligent reflexes''). These reflexes are governed by a control policy which maps states of the body to motor commands in a manner conducive to accomplishing particular goals. These control policies are distinguished from models, since they need not involve any explicit model of the problem or any form of simulation. 

This feature of human cognition has several advantages not least of these is that it allows human beings to build up a toolbox of useful control policies which can be activated under appropriate circumstances without requiring effortful thought. Further, given our use of language, we can transmit propositional knowledge to other individuals, allowing them to transform this knowledge into control policies without an extended process of deliberation or discovery by trial and error. Krakauer notes that it is almost impossible to teach other animals certain skills because we cannot use language to transmit propositional knowledge of how to perform an action---even though we do not rely on that propositional knowledge when producing the action. For example, teaching another person to juggle requires explaining a few crucial steps. Knowing these steps is sufficient to begin automatizing juggling, and it is very difficult to acquire the skill of juggling without this knowledge. 

Unlike other speakers (e.g., Olshausen), Krakauer is hesitant to invoke models when explaining intelligent behavior. Of course, there is probably \textit{some} sense in which a control policy encodes relevant information about a problem, but calling this a ``model'' risks trivializing the concept. For this reason, Krakauer argues that we should reserve talk of ``models'' for cases where we actually use a model for simulation or prediction. 

In the post-talk discussion, the extent to which models play a role in the production of intelligent behavior was a point of significant contention.  While all parties recognized the importance of both models and reflexes, there was no consensus on what sorts of  intelligent behavior require a model of the world. Another point of contention was the viability of the System 1/System 2 distinction at play in Krakauer's view (Evans \& Stanovich, 2013; Kahneman, 2011). Some argued that this distinction does not carve cognition at the joints, with many forms of cognition defying easy classification as System 1 or System 2 processes. In response, Krakauer argued that the highly selective deficits of stroke patients suggest that reflexive behaviors (at least) are compartmentalized (e.g., a patient who can sign their name, but not compose a simple sentence). 

\subsection[``All Intelligence Is Collective, but Not All Collectives Are Intelligent'' (Melanie Moses)]{``All Intelligence Is Collective, but Not All Collectives Are Intelligent''}

\begin{adjustwidth}{1cm}{1cm}
\onehalfspacing
\textit{Melanie Moses is a Professor of Computer Science at the University of New Mexico and an External Professor at the Santa Fe Institute.}
\end{adjustwidth}

\noindent Melanie Moses discussed the relationship between biological, human, and collective intelligence as well as the proper use of intelligence and intelligent systems. Moses begins by discussing biological intelligence. She argues that biological intelligence is characterized by possessing valuable environmental maps, having the ability to use such maps to increase one's fitness, and being able to change the world to make one's maps more valuable. A key aspect of these ``maps'' is that a map's value is highly contingent on the map-world relationship. For example, the value of a treasure map depends not only on how detailed that map is, but also on how effectively those details guide us to the treasure, how much treasure is there, etc. Analogously, an apple seed possesses a map of the environmental indicators that signal the best time to begin germination. Of course, the seed is highly dependent on these indicators being reliable---something that is dependent on the specific environment and learned over evolutionary time. On this view, general intelligence amounts to having many maps or having maps that remain useful across diverse environments. 

Next, Moses discusses the distinctive features of human intelligence, arguing that uniquely human intelligence ``includes understanding, meta-cognition, and a heightened ability to teach and learn'' and that this intelligence ``emerges from being conscious of our own understanding." Following Douglas Hofstadter, she proposes that central to uniquely human understanding is the fact that we have a model or map of the world \textit{and} a model of our own minds (Hofstadter, 2007). Further, the human ability to learn and teach allows us to (collectively) extend this understanding beyond our individual capacities. Cumulative cultural evolution occurs via the transmission and accumulation of cultural knowledge and institutions. 

Finally, Moses considers the distinctive features of collective intelligence. For example, groups of ants (being spread out over their environment) are able to use markers (e.g., pheromones) to create maps of the world using simple behavioral rules governing when to mark a path after a cache of resources is discovered. Further, in evolutionary time, these rules can be adapted to the specific distribution of food found in their environment (e.g., widely and evenly or concentrated and patchy). More generally, collective intelligences have a special facility for balancing explore-exploit trade-offs by having different numbers of members pursuing each objective. They also have the ability to organize into systems and react to their own organization. In addition, they often blur the boundary between self and environment, and their cooperation is often driven by difficulty in distinguishing their own actions from those of others (e.g., difficulty in linking chemical signals to particular individuals).

Turning to the use of intelligent systems, Moses highlights some potentially valuable uses for artificial intelligence. In particular, she considers the ways that human intelligence could be augmented by enhancing the distinctive virtues of each form of intelligence discussed previously. For example, perhaps we can use AI systems to improve our ability to teach and learn, to more thoroughly explore solutions to difficult problems, and enhance our ability to communicate and cooperate. We might also be able to use AI research to better understand how our own intelligence is embedded in natural and human-created systems. For example, we might use machine learning trained on human-generated data sets to discover biases in human culture and cognition. That said, Moses is also cautious about AI risks---not because she believes that AI will be radically different or vastly more intelligent than humans, but because the methods humans have already used to extend our intelligence (e.g., cultural transmission and social institutions) have already posed serious risks. Teams of scientists have built on prior culturally-transmitted understanding to both improve our quality of life and to create weapons that pose existential risks. To proceed safely, then, we should be careful to recognize the danger inherent in intelligence itself in addition to any risks posed by AI in particular.

\subsection[``The Elements of Collective Intelligence'' (Jessica Flack)]{``The Elements of Collective Intelligence''}

\begin{adjustwidth}{1cm}{1cm}
\onehalfspacing
\textit{Jessica Flack is a Professor at the Santa Fe Institute where she directs the Collective Computation Group.}
\end{adjustwidth}

\noindent Jessica Flack described the relationship between collective intelligence, collective computation, and collective pattern formation, and she outlined a framework for characterizing collective computation in biological systems. Flack begins by discussing work on collective pattern formation, which focuses on how components interacting according to simple rules give rise to complex patterns (e.g., synchronized behavior). Here, the focus is on the emergent patterns of activity. In contrast, work on collective intelligence is primarily concerned with whether the output of collective behavior solves a problem or maximizes fitness. For example, work on social insects might show how the rules governing individual ant behavior allow the collective (in aggregate) to produce an adaptive outcome. Finally, the collective computation literature blends elements of theoretical computer science and mechanics to describe how collectives engage in computation. 

Flack is primarily interested in the intersection of collective computation and collective intelligence, and her research aims at understanding the computational mechanisms by which collectives bring about adaptive outcomes. To do so, she argues, we need a suitable formal framework for discussing collective computational in biological systems. Collective computation's focus on mechanics is especially important here, as biological systems face information processing and thermodynamic constraints that must be considered in understanding and evaluating how collectives solve problems. It is also important to consider collective intelligence's focus on how evolution drives collectives to discover adaptive solutions to problems. Taken together, we need a formal framework that allows us to not only explain the mechanisms of collective computation, but also to understand how these mechanisms can be tuned or improved by evolution (or even learning) to produce adaptive outputs. 

To this end, Flack outlines seven key elements of collective intelligence: environment, input, output, task, circuit, algorithm, and termination criteria. In other words, we must specify what variables are fixed by the environment, what variables the system can respond to, what behaviors are produced in aggregate, what problem those behaviors solve, what components there are/how those components are organized, the process by which those components produce solutions, and how a system determines when a solution has been found. While this framework is necessarily quite general, it helps to focus our attention on what is required to rigorously characterize biological computation. For example, within this framework we can ask how a collective changes via learning. We might also develop statistical tools for identifying relevant components or the environmental variables to which those components are sensitive. As Flack argues, understanding the elements of collective intelligence in these ways is likely central to understanding intelligence in general, since any intelligence will ultimately be realized by many interacting parts.

\subsection[``Intelligence Needs (At Least) Two Blades'' (Mirta Galesic)]{``Intelligence Needs (At Least) Two Blades''}

\begin{adjustwidth}{1cm}{1cm}
\onehalfspacing
\textit{Mirta Galesic is a Professor at the Santa Fe Institute, External Faculty at the Complexity Science Hub in Vienna, Austria, and Associate Researcher at the Harding Center for Risk Literacy at the University of Potsdam, Germany.}
\end{adjustwidth}

\noindent Mirta Galesic challenged our current, human-centered understanding of intelligence and proposed a novel approach based on ecological intelligence. Galesic begins by outlining the deficiencies of our current picture of intelligence. For example, she notes that there are a wide variety of competing and imprecise definitions of intelligence. Further, she notes that we often focus on human intelligence as basis for understanding intelligence in general. This presents difficulties in that several animals which lack many of our cognitive abilities actually exceed human performance on cognitive tests---even tests emphasizing rational decision making. For example, birds have been shown to perform better than humans in a scenario based on the notoriously counterintuitive Monty Hall problem (Herbranson \& Schroeder, 2010). The point here is not that this bird species is actually more intelligent than human beings, but rather that we need a better notion of intelligence to properly appreciate the distinctive cognitive abilities of diverse agents.

Galesic proposes that \textit{ecological rationality} might provide a better model of intelligence than human intelligence. Ecological rationality is the degree to which an agent's toolbox of cognitive strategies and heuristics is adapted to its environment. By understanding intelligence as a kind of fit between a cognitive system and the environment, we can more completely and more rigorously characterize each item in an agent's cognitive toolbox. To do so, we can not only characterize different cognitive strategies and heuristics, but compare the performance across a range of environments. For example, rather than asking whether a particular balance between exploration and exploitation is optimal, we can consider the success of that balance (and others) across a range of values for relevant environmental variables. These variables might characterize the task (e.g., predictability, risk, or complexity) or elements of the social environment (e.g., the presence of socially-transmitted information or of agents with whom one might cooperate or compete). 

The larger picture then is one in which researchers can (i) characterize a wide variety of cognitive strategies, (ii) characterize environments along a range of relevant dimensions, and (iii) apply a measure of fit to determine what strategies are appropriate in which environments. Echoing an idea proposed earlier in the workshop by Daniel Dennett, Galesic suggests that general intelligence could be understood as a kind of meta-cognition, where we examine a problem, find a good representation of that problem, and select an appropriate cognitive tool for solving it. This is particularly relevant to the development of general AI, she argues, because we now have a large toolbox of AI strategies appropriate to narrow domains, but we have few AI strategies for selecting and applying these tools where appropriate.

\subsection[``Beyond 20th Century Notions of `an Intelligence''' (David Wolpert)]{``Beyond 20th Century Notions of `an Intelligence': Using Stacking to Combine Intelligences''}

\begin{adjustwidth}{1cm}{1cm}
\onehalfspacing
\textit{David Wolpert is a Professor at the Santa Fe Institute.}
\end{adjustwidth}

\noindent David Wolpert discussed the power of combining models to enhance predictive accuracy. While there are a host of methods for selecting the best model of some data (e.g., k-fold cross validation), there are often gains to be found by combining the predictions of multiple models. For example, we might train several models and then train a \textit{meta-model} which takes the predictions of those models as inputs (see e.g., Breiman, 1996; Yao et al., 2018). This ``stacking'' process allows the meta-model (i) to incorporate the predictions of the stacked models and (ii) to model and compensate for the pattern of biases they exhibit (Wolpert, 1992, p. 251). Wolpert highlights several concrete applications of stacking, and details how stacking might be applied to Monte Carlo methods. 

The larger view on offer, however, concerns the future of the scientific method and of artificial intelligence. For example, while stacking is very difficult to formalize in terms of Bayesian statistics, and therefore has not yet been formally justified, stacking routinely outperforms the best single models. Wolpert argues that stacking represents an improvement to the scientific method which has heretofore favored something closer to the single model approach of cross-validation. Not only should we move away from this approach, he argues, but there are likely gains to be made from stacking a greater diversity of model types. With respect to artificial intelligence research, Wolpert argues that too much emphasis has been placed on the development of intelligent individuals. On the contrary, he suggests that general and super-human intelligence are likely to arise in a web-based ecology of AI systems and that this intelligence may be hard or impossible for humans to recognize from the outside. 

During the discussion, at least two important issues emerged. First, Melanie Mitchell asked whether communities of AI systems were really different in kind from communities of humans. While Wolpert conceded that human communities are not different in kind, he warns that AI systems could operate on time scales far different than those of human activity, resulting in phenomena like the 2010 ``flash crash.'' He also emphasized that because the individual AI programs were Turing complete, the behavior of a community of them is, in a formal sense, uncomputable. As such, we cannot predict their behavior even in theory. A second important issue was whether stacking models actually helps us to understand the systems we model. A single model will, in many cases, suggest some coherent interpretation of what is going on in the target system. It is much less clear what understanding could be gained from several distinct models whose contributions to prediction are mediated by an (additional) meta-model. In response, Wolpert cited the willingness of engineers to forgo full understanding in favor of predictive accuracy as a model for future attitudes in science. 

\subsection[``Evolving Intelligent Behavior in Virtual Agents'' (Risto Miikkulainen)]{``Evolving Intelligent Behavior in Virtual Agents''}

\begin{adjustwidth}{1cm}{1cm}
\onehalfspacing
\textit{Risto Miikkulainen is a Professor of Computer Science at the University of Texas at Austin and AVP of Evolutionary Intelligence at Cognizant Technology Solutions.}
\end{adjustwidth}

\noindent Risto Miikkulainen discussed the role of constraints in the evolution of intelligent systems in natural and artificial environments. Miikkulainen begins by outlining his approach to the study of intelligence---using the evolution of neural networks to build agents whose intelligent behavior emerges as the solution to some problem the agent faces. While this still requires some elements of human design, the goal is to (increasingly) get more out of evolving these systems than is initially seeded by human engineers. 

When human or animal-like behavior is desired outcome, Miikkulainen argues, we must constrain evolution in realistic ways. For example, he discusses ``Botprize,'' a competition to design human-like opponents in a video game environment. Notably, it was not difficult to evolve AI systems that were successful at playing the game, but these AI systems did not initially exhibit human-like play. However, when realistic constraints were imposed (e.g., to accuracy, reaction time, etc.), the AI systems behaved in ways that appeared substantially more human to their human opponents (Schrum, Karpov, \& Miikkulainen, 2012). 

Of course, these AI systems were essentially controlling an in-game character rather than evolving basic ways of moving and interacting with their environment. What if the behavior we are seeking to evolve are these basic bodily actions? Again, Miikkulainen argues, constraints can help in the evolution of realistic behaviors. In this case, constraints are imposed by forcing the body \textit{and} behavior of virtual organisms to co-evolve. Rather than giving evolution a highly-capable body from the outset and allowing it to shape behavior, evolution must solve the harder problem of finding good body/behavior combinations. Another useful constraint is to place agents in environments where arms races can emerge with competitors. Again, these constraints resulted in remarkably intelligent and animal-like behaviors (Lessin, Fussell, \& Miikkulainen, 2014). 

The importance of constraints in the evolution of intelligent behavior seems likely to extend beyond virtual environments. For example, hyenas have evolved surprisingly sophisticated coordinated hunting strategies in order to steal kills from lions. Since an individual hyena would not fare well against a lion, cooperation is key to driving the lions from their kill. Computer simulations of this scenario show a plausible path from individual risk-taking behaviors to highly-coordinated attacks (Rajagopalan, Holekamp, \& Miikkulainen, 2020). The larger picture of intelligence offered by Miikkulainen is that intelligence arises from the evolutionary optimization of neural networks under a range of realistic constraints, including resource limitation, body-brain co-evolution, competitive arms races, and the need for coordination.  

\subsection[``Evolution and Engineering'' (Stephanie Forrest)]{``Evolution and Engineering''}

\begin{adjustwidth}{1cm}{1cm}
\onehalfspacing
\textit{Stephanie Forrest is a Professor in the School of Computing, Informatics and Decision Sciences Engineering (CIDSE) at Arizona State University, where she also directs the Biodesign Center for Biocomputation, Security and Society.  She is also an External Professor at the Santa Fe Institute.}
\end{adjustwidth}

\noindent Stephanie Forrest discussed the power of evolutionary algorithms in software development and AI. Forrest begins by tackling some of the central questions of the workshop. She expresses skepticism that ``intelligence'' can ever be properly defined, but notes that intelligence seems to be a ``collection of interacting capacities that involve information processing.'' Further, a striking fact is that these diverse capacities are implemented in approximately the same kind of substrate (i.e., nervous systems) and that all these capacities evolved over time by the process of natural selection. This suggests that evolution may be of particular importance in understanding and developing artificial intelligence.

While nature made use of nervous systems, Forrest argues that the artificial intelligence will most likely be implemented in software. Further, she argues that software itself is evolved.  This evolution can take different forms. Software might evolve at the level of individual programs, or at the level of software environments as they are used and modified by human communities. Forrest's research focuses on the former, a type of evolutionary computation, noting that many of today's software engineering problems are too large and complicated for human developers, and that the foresighted process of human design still falls prey to unanticipated uses, abuses, and interactions.

That said, there are many ways in which evolution in software (at least as presently practiced) falls short of biological evolution. Despite substantial success in using evolutionary algorithms to improve software and neural networks (see e.g., Liou et al., 2020), Forrest argues that biological evolution has a number of characteristics worth borrowing to advance the field of evolutionary computation. These include: open-endedness, major transitions, neutrality and drift, genotype/phenotype mappings, multi-objectivity, and co-evolution (Miikulainen \& Forrest, 2021). She suggests that many of these can be imported into evolutionary algorithms either by directly emulating biological evolution or by scaling up existing approaches. In addition to augmenting existing approaches to evolutionary computation, Forrest also discusses how the processes of evolution and engineering, which are usually thought of as distinct processes, have become increasingly entangled and indistinct.  She emphasized the promise of hybrid approaches that further blur the lines between evolutionary algorithms and conventional engineering approaches. Together, she argues, these approaches may allow us to construct genuinely intelligent AI systems.

\section{Conclusion}

During the workshop, speakers addressed a wide variety of ideas, from high-level theories about the nature of intelligence to specific experimental results. That said, there were several recurring themes. Some of these themes represent points of consensus, others disagreements, and still others areas where more research is needed. One point of broad consensus was that intelligence can take multiple forms and that different definitions of intelligence are appropriate to different areas of inquiry. Several speakers drew distinctions between different forms of intelligence in order to emphasize the difficulty of bestowing a particular form of intelligence on artificial systems. 

For example, Melanie Mitchell argued that we often underestimate the difficulty of giving AI systems general intelligence by assuming that progress on domain-specific intelligence is progress towards general AI. John Krakauer, in turn, emphasized that we have no idea what higher-level thinking really is despite the fact that we have made considerable progress in understanding more reflexive forms of intelligence. Other speakers drew distinctions between different forms of intelligence in order to explain particular kinds of intelligent behavior. Ricard Sol\'{e} described the behavior of ant colonies as arising from a distinctive ``liquid brain'' architecture, and David Krakauer emphasized the ``learned incomprehension'' which underlies high-level performance in competitive puzzle solving.

One area of contention was how best to understand the distinction (if there is one) between effortful System 2-style cognition and automatic System 1-style cognition. In particular, John Krakauer's talk elicited significant disagreement about how to approach this distinction. For example, Alison Gopnik argued that the distinction is not particularly useful in developmental psychology and that a more nuanced view is required. Others (e.g., Bruno Olshausen) raised concerns about what kinds of intelligent behavior could really be accounted for by the kind of automatized algorithms/control policies described by David and John Krakauer. This dispute is closely linked to Daniel Dennett's discussion of ``competence without comprehension'' and how human problem-solving exceeds this modest standard.  Of course, in both cases, further research is required to determine exactly which mechanisms do (or perhaps could) underlie which intelligent behaviors. 

Another broad area of consensus was the importance of social networks in understanding intelligence. This took two forms. First, there was an emphasis on social groups as providing a kind of scaffolding for learning. Alison Gopnik argued that human beings have evolved distinctive strategies (e.g., extended childhood and menopause) that facilitate the social transmission of knowledge. David Krakauer described how knowledge of novel techniques is disseminated to individuals in communities of competitive game players. Thalia Wheatley emphasized how social connections can help to enhance our intelligence by connecting us with diverse perspectives. 

Second, several speakers emphasized intelligence \textit{itself} as a social or collective phenomenon. Melanie Moses discussed biological, human, and collective intelligence--describing how collective intelligence is a powerful tool for extending both biological and human intelligence. As noted earlier, Ricard Sol\'{e} described the special architecture underlying collective intelligence in ants. Jessica Flack took a more fundamental approach, considering how we might (in general) model computation and intelligence as arising from the networks of interacting components. 

Another area of convergence was the idea that general intelligence involves a kind of higher-order cognition. Daniel Dennett, for example, discussed how intelligence involves knowing how to deploy heuristics appropriate to a specific problems. Mirta Galesic echoed this idea arguing that organisms possess a toolbox of adaptive strategies and heuristics and that intelligence might be viewed as the ability to construct a representation of a problem that allows one to select the right tool for the job. Melanie Moses argued that a distinctive feature of human intelligence is that we not only model our environment, but we model ourselves and our own cognitive processes. John Krakauer argued that there is a central role for System 2 cognition in discovering or learning novel skills that will (ultimately) be added to our set of ``intelligent reflexes'' and deployed where appropriate. At the same time, there was less consensus about whether and to what degree conscious comprehension is required to characterize a cognitive process as intelligent.

The speakers also generally recognized the importance of evolution in shaping intelligence. For some (e.g., Mirta Galesic), the constraints imposed by evolution are integral to understanding intelligence which (one may argue) is difficult to rigorously define without understanding an agent's environment and limitations. For others (e.g., Risto Miikkulainen), the constraints imposed by evolution are a key driver of intelligent behavior and should be replicated when evolving artificial agents. Still others (e.g., Stephanie Forrest) emphasized the distinctive features of evolution as a design process, arguing that it has the potential to replace or augment existing approaches to software development \textit{if} it can be tweaked and scaled so as to better emulate real-world evolutionary processes. More broadly, there was a general recognition that evolution is the only known process that has resulted in robust general intelligence and that this cannot be ignored as we approach AI. 

This leads to another theme that recurred several times during the workshop: the importance of looking to diverse forms of natural intelligence when building or studying artificially intelligent systems.. This extends beyond the role of evolution or evolutionary constraints. As noted earlier, Mirta Galesic argued that the notion of ecological intelligence (i.e., intelligence as a kind of adaptation to one's environment) can help us to more rigorously define intelligence by appealing to specific ecological contexts, contexts where the constraints and success conditions can be more clearly specified. Alison Gopnik argued that adaptations to human cognition and life history suggest different modes of intelligence that, when realized in human populations, help to solve the problem of balancing exploration and exploitation. More generally, there seemed to be a sense that by examining the variety of ways in which natural intelligence differs from existing artificial intelligence, we might identify important strategies for improving AI systems.

In summary, the workshop brought together experts from a wide range of fields to discuss and develop an interdisciplinary taxonomy of intelligence. Despite their diverse backgrounds, the speakers converged on a number of major themes---with only a few pointed disagreements. Crucially, the speakers addressed these themes by appeal to specific work from their own disciplines, introducing new and important research (and researchers) to experts in other fields. Hence, the workshop both (i) furthered a common understanding of how to approach intelligence in natural and artificial systems and (ii) provided concrete starting points for interdisciplinary research and collaboration at the cutting edge of intelligence research. 

\subsubsection*{Acknowledgments}
This material is based upon work supported by the National Science Foundation under Grant No.\ 2020103.  Any opinions, findings, and conclusions or recommendations expressed in this material are those of the author and do not necessarily reflect the views of the National Science Foundation. This work was also supported by the Santa Fe Institute. The authors of this report are especially indebted to the speakers for contributing their time and effort to the workshop and to student assistants Julie Hayes and Abigail Pribosova for taking careful notes during the workshop. 

\newpage

\section{References}

\begin{flushleft}

\begin{list}{}
{\leftmargin=1em \itemindent=-1em \itemsep=-.4em}

\item Breiman, L. (1996). Stacked regressions. \textit{Machine Learning}, \textit{24}, 49-64.

\item Dennett, D.C. (2008). A route to intelligence: Oversimplify and self-monitor. \url{https://ase.tufts.edu/cogstud/dennett/papers/oversimplify.pdf}

\item Dallenbach, K.M. (1951). A puzzle-picture with a new principle of concealment. \textit{The American Journal of Psychology}, \textit{64}(3), 431-433.

\item Evans, J. (2020). Social computing unhinged. \textit{Journal of Social Computing}, \textit{1}(1), 1-13.

\item Evans, J. S. B., \& Stanovich, K. E. (2013). Dual-process theories of higher cognition: Advancing the debate. \textit{ Perspectives on psychological science}, \textit{8}(3), 223-241.

\item Gopnik, A., Fankenhuis, W.E., \& Tomasello, M. (2020). Introduction to special issue: Life history and learning: How childhood, caregiving and old age shape cognition and culture in humans and other animals. \textit{Philosophical Transactions of the Royal Society B}, \textit{375}(1803), 1-6.

\item Herbranson, W. T., \& Schroeder, J. (2010). Are birds smarter than mathematicians? Pigeons (columba livia) perform optimally on a version of the Monty Hall Dilemma. \textit{Journal of Comparative Psychology}, \textit{124}(1), 1-13.

\item Hofstadter, D. (2007). \textit{I am a strange loop}. New York, NY: Basic Books. 

\item Kahneman, D. (2011). \textit{Thinking fast and slow}. New York, NY: Farrar, Straus, and Giroux.

\item Hill, D. E. (1979). Orientation by jumping spiders of the genus Phidippus (Araneae: Salticidae) during the pursuit of prey. \textit{Behavioral Ecology and Sociobiology}, \textit{5}(3), 301-322.

\item Klein, A. (2019). The curious case of the decapitated frog: On
experiment and philosophy. \textit{British Journal for the History of Philosophy}, \textit{26}(4), 1-28.

\item Krakauer, J.W. (2019). The intelligent reflex. \textit{Philosophical Psychology}, 32(5), 822-830.

\item Lessin, D., Fussell, D., \& Miikkulainen, R. (2014). Adopting morphology to multiple tasks in evolved virtual creatures. In \textit{Artificial Life Conference Proceedings 14}. Cambridge, MA: MIT Press. 247-254.

\item Liou, J. Y., Wang, X., Forrest, S., \& Wu, C. J. (2020). GEVO: GPU code optimization using evolutionary computation. \textit{ACM Transactions on Architecture and Code Optimization}, \textit{17}(4), 1-28.

\item McDermott, D. (1976). Artificial intelligence meets natural stupidity. \textit{ACM SIGART Bulletin}, 57. 4-9.

\item McGeer, V. (2007). The regulative dimension of folk psychology. In D.D. Hutto and M. Ratcliffe (eds.) \textit{Folk psychology re-assessed}. Dordrecht, Netherlands: Springer. 137-156.

\item Miikkulainen, R., \& Forrest, S. (2021). A biological perspective on evolutionary computation. \textit{Nature Machine Intelligence}, \textit{3}(1), 9-15.

\item Minskey, M. (1988). \textit{Society of mind}. New York, NY: Simon and Schuster. 

\item Moravec, H. (1988). \textit{Mind children: The future of robot and human intelligence}. Cambridge, MA: Harvard University Press.  

\item Parkinson, C., Kleinbaum, A., \& Wheatley, T. (2018). Similar neural responses predict friendship. \textit{Nature Communications}, \textit{9}:332, 1-14.

\item Parkinson, C., Kleinbaum, A., \& Wheatley, T. (2017). Spontaneous neural encoding of social network position. \textit{Nature Human Behavior}, \textit{1}:72, 1-7.

\item Rajagopalan, P., Holekamp, K. \& Miikkulainen, R. (2020). Evolutions of complex coordinated behavior. In \textit{2020 IEEE Congress on Evolutionary Computation}. 1-8. 

\item Schrum, J., Karpov, I., \& Miikkulainen, R. (2012). Humanlike combat behavior via multiobjective neuroevolution. In P. Hingston (Ed.) \textit{Believable Bots}. New York, NY: Springer. 119-150.

\item Seelig, J. D., \& Jayaraman, V. (2015). Neural dynamics for landmark orientation and angular path integration. \textit{Nature}, \textit{521}(7551), 186-191.

\item Sievers, B., Welker, C., Hasson, U., Kleinbaum, A.M., Wheatley, T. (Preprint). How consensus-building conversation changes our minds and aligns our brains. \url{https://psyarxiv.com/562z7/}.

\item Sourati, J., \& Evans, J. (2021). Accelerating science with human versus alien artificial intelligences. \url{https://arxiv.org/pdf/2104.05188.pdf}.  

\item Tarsitano, M. \& Jackson, R. (1997). Araneophagic jumping spiders discriminate between detour routes that do and do not lead to prey. \textit{Animal Behavior}, 53, 257-266.

\item Wood, A., Kleinbaum, A.M., Wheatley, T. (2019). Cultural Diversity Broadens Social Networks. \url{https://psyarxiv.com/qvthk/}.

\item Wolpert, D. (1992). Stacked generalization. \textit{Neural Networks}, \textit{5}(2), 241-259.

\item Yao, Y., Vehtari, A., Simpson, D., and Gelman, A. (2018). Using stacking to average Bayesian predictive distributions. \textit{Bayesian Analysis}, \textit{13}(3), 917-1007.

\item Zhang, K. (1996). Representation of spatial orientation by the intrinsic dynamics of the head-direction cell ensemble: A theory. \textit{Journal of Neuroscience}, \textit{16}(6), 2112-2126.

\end{list}
\end{flushleft}

\end{document}